%% file: main.tex
\documentclass[letterpaper, 10 pt, conference]{ieeeconf}  
\IEEEoverridecommandlockouts                              
\usepackage{packages}
\usepackage{multirow}
\usepackage{amsmath}

\title{\LARGE\bf Multi-Modal Multi-Agent Optimization for LIMMS, \\ A Modular Robotics Approach to Delivery Automation}
\author{Xuan Lin$^{1}$, Gabriel I.~Fernandez$^{1}$, Yeting Liu$^{1}$, Taoyuanmin Zhu$^{1}$, Yuki Shirai$^{1}$, and Dennis Hong$^{1}$
\thanks{
$^{1}$Xuan Lin, Gabriel I.~Fernandez, Yeting Liu, Taoyuanmin Zhu, Yuki Shirai and Dennis Hong are with RoMeLa (the Robotics and Mechanisms Laboratory) at the University of California, Los Angeles, USA. {\tt\small \{maynight, gabriel808, liu1995, tymzhu, yukishirai4869, dennishong\}@ucla.edu}}
}

\begin{document}
\maketitle
\thispagestyle{empty}
\pagestyle{empty}

\begin{abstract}
In this paper we present a motion planner for LIMMS, a modular multi-agent, multi-modal package delivery platform. A single LIMMS unit is a robot that can operate as an arm or leg depending on how and what it is attached to, e.g., a manipulator when it is anchored to walls within a delivery vehicle or a quadruped robot when 4 are attached to a box. Coordinating amongst multiple LIMMS, when each one can take on vastly different roles, can quickly become complex. For such a planning problem we first compose the necessary logic and constraints. The formulation is then solved for skill exploration and can be implemented on hardware after refinement.
To solve this optimization problem we use alternating direction method of multipliers (ADMM).
The proposed planner is experimented under various scenarios which shows the capability of LIMMS to enter into different modes or combinations of them to achieve their goal of moving shipping boxes.



\end{abstract}


\input{Sections/S1_introduction}
\input{Sections/S2_Hardware}
\input{Sections/S3_problem}

\input{Sections/S4_solving_with_ADMM}
\input{Sections/S5_Results}
\input{Sections/S6_Conclusion}

\section*{Acknowledgements}
The authors would like to thank LG Electronics for sponsoring this research and giving useful feedback. We would also like to thank other lab members of RoMeLa who have contributed to LIMMS development: Justin Quan, Colin Togashi, and Samuel Gessow.
\vspace{-0.1cm}
{
\bibliographystyle{IEEEtran}
\bibliography{references}
}


\end{document}

%% file: Sections/S1_introduction.tex
\section{Introduction}
\label{sec:introduction}
\vspace{-0.1cm}
Modular self-reconfigurable robot systems have attracted the attention of researchers~\cite{yim2007modular, liu2019reconfiguration, zhao2018design}.
The ability to transform relatively simple robot modules to realize various forms and functions can be applied in areas such as aerial or space robotics \cite{stieber1999overview, cruijssen2014european} and more recently last-mile delivery \cite{zhu2022feasibility}. Self-reconfigurable robots present challenging mechanical design, motion planning, and control problems. As the number of robots scales up, issues arise in communication, coordination, and decision making partially because it can no longer be assumed to have accurate global information.

\begin{figure}[t!]
		\centering
		\includegraphics[width=0.85\linewidth]{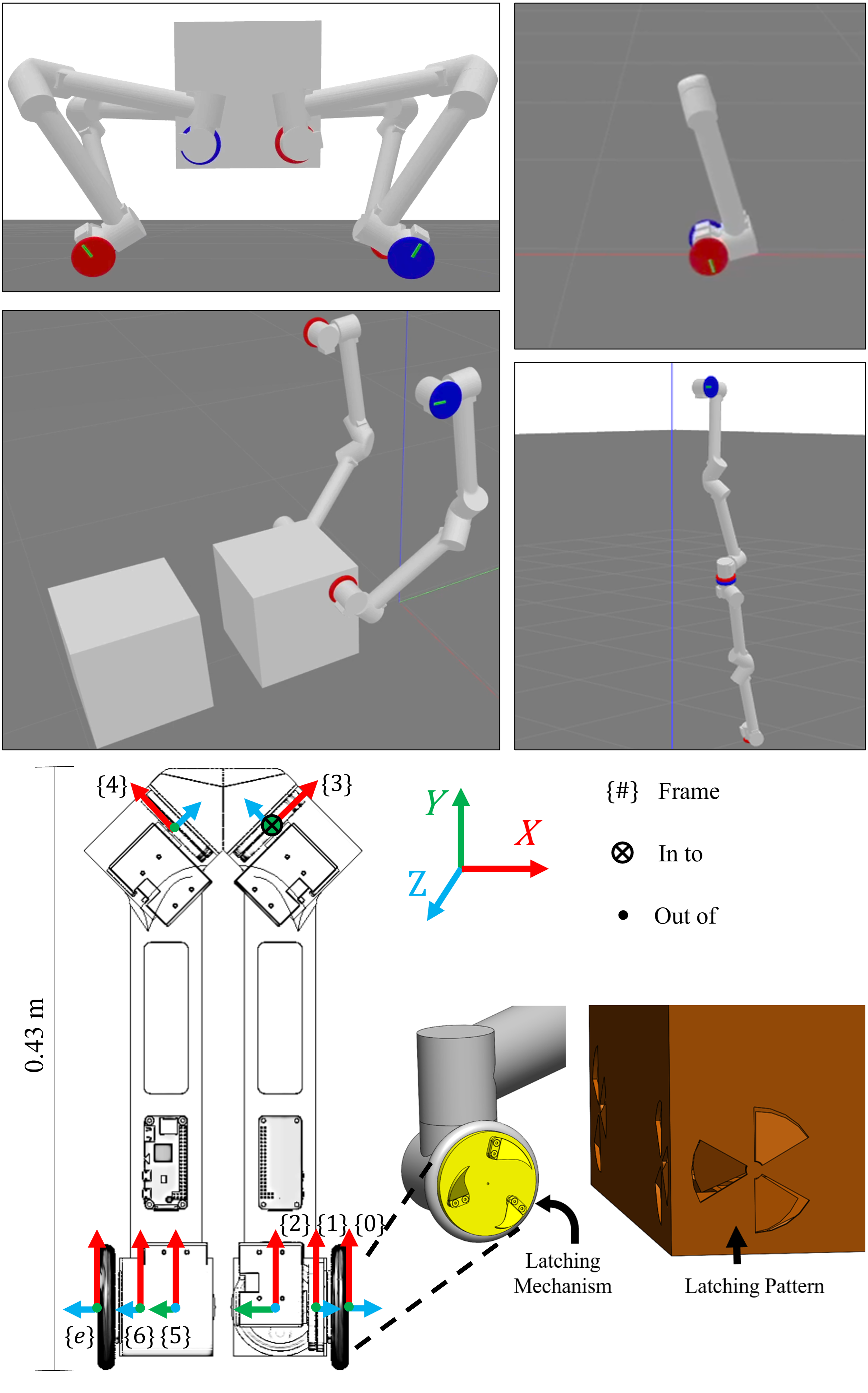}
		\caption {\textit{Top} shows the various operational modes for LIMMS which will be considered for the optimization formulation. \textit{Upper Left}: 4 LIMMS attached to a box in quadruped mode. \textit{Upper Right}: 1 LIMMS self-balancing to move. This is called free mode. \textit{Lower Left}: 2 LIMMS anchored to surfaces in manipulator mode. \textit{Lower Right}: LIMMS attached to each other. This can be used in both quadruped and manipulator mode. \textit{Bottom} shows the joint axis and the latching mechnanism to be integrated at the end effector.}
		\label{fig:main}
\end{figure}

In recent work, we introduced LIMMS (Latching  Intelligent  Modular  Mobility System) as a modular approach to last-mile delivery \cite{zhu2022feasibility}, shown in Fig.~\ref{fig:main}. LIMMS system is composed of individual modules that resemble a 6 degree-of-freedom (DoF) arm but with wheels and a latching mechanism at both ends. As such either side of LIMMS can act as the base depending on the need. In the case of last-mile delivery, within the delivery truck, it is assumed that the surface, as well as boxes, have anchor points LIMMS can attach the latch to. By attaching one end to the walls within the vehicle LIMMS can carry and move boxes by latching its free end to them like other manipulators. To actually transport the box to the recipient's door, four LIMMS can attach to a box and use it as its body to walk itself there. Finally, to return LIMMS can self-balance and wheel itself back.
LIMMS introduces a challenging motion planning problem with each LIMMS having multiple DoF. This requires kinematic constraints which were rarely done in previous works \cite{liu2019distributed, gilpin2008miche}.
Moreover, each LIMMS operate in different modes: a wheeled robot, a leg, or an arm. The delivery package can also be transported in various ways: manipulation or locomotion via acting as the body for a quadruped robot. This introduces complicated mixing of discrete decisions and continuous constraints.

Various approaches have been explored to resolve the reconfiguration planning problem, such as graph search methods \cite{liu2019distributed}, reinforcement learning \cite{zhang2021reinforcement}, or optimization-based approaches \cite{gandhi2020self}. We adopt optimization-based methods for LIMMS. Among the optimization-based methods, mixed-integer programs (MIP) are useful for discrete decision-making within multi-agent systems \cite{schouwenaars2001mixed, kataoka2020multi}. However, the LIMMS motion planning problem includes nonlinear constraints such as kinematics. Therefore, the problem becomes a mixed-integer nonlinear (non-convex) program (MINLP). MINLPs are known to be computationally difficult when the problem scale is large. Aside from directly applying commercialized solvers, e.g., SCIP, there are generally two approaches that transform the MINLP problems into MIPs using linear approximations of the nonlinear constraints, or into nonlinear programs (NLPs) using complementary constraints to replace binary variables. Unfortunately, as the problem scales up, nonlinear constraints yield a large amount of piece-wise linear approximations making the problem very slow \cite{lin2021designing}, and complementary constraints tend to cause infeasibility without a good initial guess \cite{lin2022reduce}.

In this paper, we implemented the alternating direction method of multipliers (ADMM) to solve the MINLP problem, inspired by \cite{aydinoglu2021real, shirai2022simultaneous}. ADMM decomposes the problem into two sub-problems: an MIP problem and an NLP problem. By decomposition, the problem scale of both MIP and NLP is reduced and becomes more tractable. The logic constraints about the connections of LIMMS are formulated into MIP while the nonlinear kinematics constraints are formulated into NLP. We establish general rules for such a system to operate under and use ADMM to explore possible approaches to resolve the given scenario, showing the feasibility of our proposed method as well as the richness of the system. To summarize, our contributions are as follows:



\begin{enumerate}
    \item Solved a unique type of problem that incorporates multi-agent locomotion and manipulation.
    \item Proposed a formulation that can be generalized to solve such types of multi-agent motion planning problems.
    \item Solved the proposed formulation with ADMM and demonstrated it on hardware.
\end{enumerate}

%% file: Sections/S2_Hardware.tex
\section{System Description}
\label{sec:Hardware}
\vspace{-0.1cm}
 


\textit{Hardware.} To gain an intuition of the constraints in the optimization formulation, the physical limitations and design of the first LIMMS hardware prototype are detailed in this section and depicted in Fig.~\ref{fig:main}. LIMMS is a symmetric 6 DoF robot. At full length, a single LIMMS unit stretches to about 0.75 m and weighs roughly 4.14 kg including batteries. From preliminary tests in simulation with a target package payload of 2 kg, we determined off-the-shelf Dynamixel motors from ROBOTIS were sufficient for the prototype. With our custom gearbox, it reaches a peak velocity of 2 rad/s and a peak torque of 31 Nm. For interested readers previous recent work in \cite{zhu2022feasibility} goes further in-depth on the overall design.


One of the crucial aspects of LIMMS is its latching mechanism, which allows it to attach either end to anchor points or itself. Latching will be used frequently since it is how LIMMS transitions to different modes and grab objects for manipulation. The current prototype proposed in \cite{latching2022} consists of a radially symmetric multi-blade design. Fig.~\ref{fig:main} (bottom right) depicts the latch prototype. By the geometry of the blade and the anchor point hole pattern, the latch mechanically self-aligns when it rotates. 

The mechanism is designed to maximize robustness to misalignment in position and orientation about its center axis, see \cite{latching2022}. 
This has a dual effect of easing control effort and allowing for the latches to pull the box or itself into the desired position and orientation without being fully positioned. If LIMMS is kinematically constrained and cannot fully reach the target position and orientation, there is slack created by the latch's mechanical design. In this sense when latching to an anchor point, the alignment does not need to strictly satisfy its constraint. 

 


\textit{Modes of Operation.} Through latching, LIMMS can enter many different modes to complete its tasks. For our proposed planner we only consider three different modes as these will be sufficient for delivering packages. However, LIMMS has the potential for many other modes, some of which are described in \cite{zhu2022feasibility}. Fig.~\ref{fig:main} depicts the 3 modes considered in our optimization formulation: 4 LIMMS attached to a box are in quadruped mode, LIMMS attached to walls or surfaces are in manipulator mode, a single LIMMS which wheels around or move like a snake is in free mode. The last sub-figure in the top half, lower right shows two LIMMS attached to the end of each other. This can be viewed as a sub-skill accessible to all modes. 

%% file: Sections/S3_problem.tex
\section{Problem Formulation}
\label{sec:problem}
In this section, we demonstrate the optimization formulation for LIMMS motion planning. The objective function of the optimization problem is to minimize the distance from the box center position to the target box position, i.e. deliver the box to the goal.
The constraints consist of two parts: logic, which is formulated into constraints using integer variables, and kinematics or dynamics, which is formulated into linear or nonlinear equations of motion. As a result, the problem is an MINLP, which will be separated into an MIP and NLP to be then solved with ADMM. Assume there are $B$ boxes and $L$ LIMMS. Each box has $S_{B}=4$ anchor points at the center of each face. The optimization is run from $t=1,...,T$ time steps. We use upper case letters to indicate constants such as the number of boxes $B$, the number of anchor points $S_{B}$, and use the lower case letters to index the quantities such as $b=1,...,B$, $s_{b}=1,...,S_{B}$. $i$ is used to index the binary variables. Upper case letters are also used as variable names. For example, $z_{B,i}$ are binary variables associated with boxes. We state the assumptions made for this formulation:

\begin{enumerate}
    \item The box does not rotate and the momentum is assumed to be balanced. This simplifies the dynamics constraints. In practice, this minimizes the damage to the contents in boxes during shipping.
    \item Multi-body dynamics of LIMMS are not enforced. This is to simplify the constraints.
\end{enumerate}

\begin{table}[t!]
\centering
\setlength{\tabcolsep}{2.5pt}
\renewcommand{\arraystretch}{1.335}

\begin{tabular}{clcl}
\hline
& \multicolumn{1}{l}{Var} & \multicolumn{1}{c}{Dim} & \multicolumn{1}{l}{Description}\\
\hline

\cellcolor{blue!10} & $z_{B,i}$ & $[B, T]$ & \begin{tabular}[c]{@{}l@{}}Mode for box $b$ at time $t$. $i=\{1: \text{stable object}$, \\ $2: \text{free object}$, $3: \text{manipulated}$, $4: \text{quadruped} \}$\end{tabular}\\ \cline{2-4}

\cellcolor{blue!10} & $z_{L,i}$ & $[L, T]$ & \begin{tabular}[c]{@{}l@{}}Mode for limb $l$ at time $t$. $i=\{1: \text{free}$, \\ $2: \text{arm}$, $3: \text{add arm}$ $4: \text{leg}$, $5: \text{add leg} \}$\end{tabular} \\ \cline{2-4}

\cellcolor{blue!10} & $z_{S,i}$ & $[B, S_{B}, L, T]$ & \begin{tabular}[c]{@{}l@{}}Mode of connection for anchor point $s_{b}$ on box $b$ \\ to limb $l$ at time $t$\\ $i=\{1: \text{empty}$, $2: \text{to arm}$, $3: \text{to leg} \}$\end{tabular}\\ \cline{2-4}

\cellcolor{blue!10} & $z_{W,i}$ & $[S_{w}, L, T]$ & \begin{tabular}[c]{@{}l@{}}Mode of connection for anchor point $s_{w}$ on wall \\ to limb $l$ at time $t$ $i=\{1: \text{empty}$, $2: \text{to arm} \}$\end{tabular}\\ \cline{2-4}

\cellcolor{blue!10} & $z_{Ac,i}$ & $[L_{pr}, L_{po}, T]$ & \begin{tabular}[c]{@{}l@{}}Mode for connection s.t. limb $l_{po}$ \\ connects as an additional limb to $l_{pr}$\\ $i=\{1: \text{not connected}$, $2: \text{connected} \}$\end{tabular} \\ \cline{2-4}

\cellcolor{blue!10} & $z_{Lc,i}$ & $[L_{pr}, L_{po}, T]$ & \begin{tabular}[c]{@{}l@{}}Mode for connection s.t. limb $l_{po}$\\ connects as an additional leg to $l_{pr}$\\ $i=\{1: \text{not connected}$, $2: \text{connected} \}$\end{tabular}\\ \cline{2-4}





\cellcolor{blue!10} & $\textbf{p}_{B}$ & $[B, T]$ & Position of center of box $b$ at time $t$ \\ \cline{2-4}

\cellcolor{blue!10} & $\textbf{R}_{B}$ & $[B, T]$ & \begin{tabular}[c]{@{}l@{}}Orientation of box $b$ \end{tabular} \\ \cline{2-4}

\cellcolor{blue!10} & $\textbf{c}_{B}$ & $[B, C, T]$ &  Position of corner $c$ for box $b$ at time $t$ \\ \cline{2-4}

\cellcolor{blue!10} & $\textbf{p}_{L}$ & $[J, L, T]$ & Position of joint $j$ of limb $l$ at time $t$ \\ \cline{2-4}

\cellcolor{blue!10} & $\textbf{R}_{L}$ & $[J, L, T]$ & \begin{tabular}[c]{@{}l@{}}Rotation matrix of joint $j$ of limb $l$ \end{tabular} \\ \cline{2-4}

\cellcolor{blue!10} & $\textbf{f}_{L}$ & $[B, S, L, T]$ & \begin{tabular}[c]{@{}l@{}}Contact force at anchor point $s$ of box $b$\\ from limb $l$ at time $t$\end{tabular} \\ \cline{2-4}




\cellcolor{blue!10} & $\textbf{a}$ & $[L_{1}, L_{2}, T]$ & \begin{tabular}[c]{@{}l@{}}Normal vector of separating plane for $l_{1}$ and $l_{2}$\end{tabular}\\ \cline{2-4}

\parbox[t]{2mm}{\multirow{-20}{*}{\cellcolor{blue!10}\rotatebox[origin=c]{90}{Continuous}}} & $b$ & $[L_{1}, L_{2}, T]$ & \begin{tabular}[c]{@{}l@{}}Offset of separating plane for $l_{1}$ and $l_{2}$\end{tabular} \\ \hline
\end{tabular}
\caption{Table of optimization variables}
\label{tab:variables}
\end{table}

All binary and continuous variables are summarized in Table~\ref{tab:variables} except those that pertain to enforcing collision avoidance with the environment for simplicity (i.e. binary variables $\delta_{Ba, i}$, $\delta_{Bg, i}$, $\delta_{La, i}$, $\delta_{Lg, i}$, and the corresponding $\lambda \in [0, 1]$ variables for convex combination). Note \textit{pr} is short for previous, and \textit{po} is short for post.

\subsection{Integral Logic Constraints}
\textit{Mode for Boxes}. We define 4 modes for each box represented by 4 binary variables: $z_{Bi}[b, t], i=1,...,4$ for the mode of box $b$ at $t$. Mode 1 is stable object mode, where the box is supported by the ground. 
Mode 2 is free object mode where the box is in the air subject to gravity. Mode 3 is manipulated object mode where the box is connected to a manipulator. Mode 4 is quadruped mode where the box is used as a robot body. We currently only allow quadruped robot for walking. This can be relaxed to incorporate more solutions such as simultaneously bipedal walking while manipulating boxes as in \cite{hooks2020alphred}. At each $t$, a box is subject to 1 mode, such that: $\sum_{i=1}^{4} z_{Bi}[b, t] = 1  \ \forall b, \ \forall t$.


\textit{Mode for LIMMS}. We define 5 modes for each LIMMS represented by 5 binary variables: $z_{Li}[l, t],\text{where } i=1,...,5$ indicating the mode of LIMMS $l$ at $t$. Mode 1 is free (wheeled) mode, where the corresponding LIMMS unit moves on the ground like a Segway robot. Mode 2 is manipulation mode, where LIMMS connects to one connection site on the wall and may connect to one box to manipulate it. Mode 3 is add arm mode, where a LIMMS can connect to another LIMMS to extend the length of the arm for a larger workspace. Mode 4 is leg mode, where LIMMS connects to a box and serves as a leg. Mode 5 is add leg mode, where it can connect to another LIMMS 
to extend the length of the leg similar to mode 3. At any time step, LIMMS can only be in one mode, such that: $\sum_{i=1}^{5} z_{Li}[l, t] = 1 \ \forall l, \forall t$.


\begin{table*}[]
\centering
\setlength{\tabcolsep}{2.5pt}
\renewcommand{\arraystretch}{1.335}
\caption{Table of logic rules}
\begin{tabular}{clll}
\hline
& \multicolumn{1}{l}{\#} & \multicolumn{1}{c}{Logic Rule Description} & \multicolumn{1}{c}{Mathematical Formulation}\\
\hline

\cellcolor{blue!10} & 1 & Box $b$ in quadruped mode, all 4 $s_{b}$ is connected to leg mode LIMMS. & $z_{B,4}[b, t] = 1 \implies \sum_{l} z_{S,3}[b, s_{b}, l, t]=1 \ \forall s_{b}$\\ \cline{2-4}

\cellcolor{blue!10} & 2 & \begin{tabular}[c]{@{}l@{}}Anchor point $s_{b}$ on box $b$ in leg mode, box $b$ is in quadruped mode.\end{tabular} & $z_{S,3}[b, s_{b}, l, t]=1 \ \exists l, \ \exists S_{B} \implies z_{B,4}[b,t]=1$ \\ \cline{2-4}

\cellcolor{blue!10} & 3 & \begin{tabular}[c]{@{}l@{}}Anchor point $s_{b}$ is connected to LIMMS $l$ as a leg, $l$ is in leg mode.\end{tabular} & $z_{S,3}[b, s_{b}, l, t]=1 \implies z_{L,4}[l, t] = 1$ \\ \cline{2-4}

\parbox[t]{2mm}{\multirow{-4}{*}{\cellcolor{blue!10}\rotatebox[origin=c]{90}{Quadruped}}} & 4 & \begin{tabular}[c]{@{}l@{}}LIMMS $l$ in leg mode, it's connected as a leg to one anchor point.\end{tabular} & $z_{L,4}[l, t]=1 \implies \sum_{s} \sum_{b} z_{S,3}[b, s_{b}, l, t] = 1$ \\ \hline

\cellcolor{blue!10} & 5 & $b$ in manipulated mode, at least 1 $s_b$ is connected to 1 arm mode $l$. & $z_{B,3}[b, t]=1 \implies \sum_{l} \sum_{s_{b}} z_{S,2}[b, s_{b}, l, t] \geq 1$ \\ \cline{2-4}

\cellcolor{blue!10} & 6 & $s_{b}$ is connected to $l$ in arm mode, $l$ is in arm or add arm mode. & $z_{S,2}[b, s_{b}, l, t]=1 \implies z_{L,2}{[}l, t{]}=1$ or $z_{L,3}{[}l, t{]}=1$\\ \cline{2-4}

\cellcolor{blue!10} & 7 & $l$ is in arm mode, $l$ is connected to one $s_w$ on the wall or ground. & $z_{L,2}[l, t]=1 \implies \sum_{s_{w}} z_{W,2}[s_{w}, l, t]=1$ \\ \cline{2-4}

\cellcolor{blue!10} & 8 & $s_w$ on wall or ground, $s_{w}$ is connected to $l$, $l$ is in arm mode. & $z_{W,2}[s_{w}, l, t]=1 \implies z_{L,2}[1,t]=1$ \\ \cline{2-4}

\cellcolor{blue!10} & 9 & LIMMS $l$ is in add arm mode, it's connected to one other LIMMS. & $z_{L,3}[l,t]=1 \implies \sum_{l_{pr}} z_{Ac}[l_{pr}, l, t] = 1$ \\ \cline{2-4}

\cellcolor{blue!10} & 10 & \begin{tabular}[c]{@{}l@{}}LIMMS $l_{pr}$ is connected to $l_{po}$, $l_{pr}$ is in arm or add arm mode,\\ $l_{po}$ is in add arm mode.\end{tabular} & \rule{0pt}{4.5ex}  \begin{tabular}[c]{@{}l@{}}$z_{Ac}[l_{pr}, l_{po}, t] = 1 \implies \begin{cases} &z_{L,2}[l_{pr}, t] = 1\\ &z_{L,3}[l_{po}, t] = 1 \end{cases}$\end{tabular} \rule[-3ex]{0pt}{0pt} \\ \cline{2-4}

\parbox[t]{2mm}{\multirow{-8}{*}{\cellcolor{blue!10}\rotatebox[origin=c]{90}{Manipulation}}} & 11 & \begin{tabular}[c]{@{}l@{}}LIMMS $l_{pr}$ is connected to $l_{po}$, $l_{pr}$ cannot connect to any box.\end{tabular} & $z_{Ac}[l_{pr}, l_{po}, t] = 1 \implies z_{S,2}[b, s_{b}, l_{pr}, t] = 0$ \\ \hline

\parbox[t]{2mm}{\multirow{-1.1}{*}{\cellcolor{blue!10}\rotatebox[origin=c]{90}{Free}}} & 12 & \rule{0pt}{3ex} \begin{tabular}[c]{@{}l@{}}Box $b$ is in stable or free object mode, all anchor points $s_{b}$ are empty.\end{tabular} & $z_{B,1}[b, t]=1 \ or \ z_{B,2}[b, t]=1 \implies z_{S,1}[b, s_{b}, l, t] = 1 \ \forall s_{b} \ \forall l$ \rule[-1.5ex]{0pt}{0pt} \\ \hline

\end{tabular}
\label{tab:logic}
\end{table*}

\textit{Mode for Anchor Points}. Each box has 4 anchor points on each side face. We define 3 modes for each anchor point by 3 binary variables $z_{Si}[b, s_{b}, l, t],\text{where } i=1,...,3$, denoting the connection mode for anchor point $s_{b}$ on $b$ to $l$ at $t$. Mode 1 is empty mode, where $s_{b}$ on $b$ is not connected to $l$. Mode 2 is arm mode where $l$ connects to $s_{b}$ as an arm. Mode 3 is leg mode where $l$ connects to $s_{b}$ as a leg. Their summation has to be 1 at each time step: $\sum_{i=1}^{3} z_{Si}[b, s_{b}, l, t] = 1 \ \forall b, \ \forall s_{b}, \ \forall l, \ \forall t$.
In addition, at each time $s_{b}$ can connect to no more than 1 LIMMS, while a given LIMMS can connect to no more than 1 $s_{b}$ at its base point. This introduces two more constraints: $\sum_{b} \sum_{s_{b}} z_{Si}[b, s_{b}, l, t] \leq 1 \ \forall l, \ \forall t$
 and $\sum_{l} z_{Si}[b, s_{b}, l, t] \leq 1 \ \forall b, \ \forall s_{b}, \ \forall t$.
$s_{b}$ on $b$ is associated with a physical connection. If latching is enforced, the position and orientation of the base of LIMMS is constrained:

\begin{equation}
\ignorespacesafterend
    \begin{aligned}
    & \textbf{p}_{L}[j=0, l, t] = \textbf{p}_{B}[b, t] + \textbf{R}_{B}[b, t]\textbf{o}[s_b] \\
    & \textbf{R}_{L}[j=0, l, t] = \textbf{R}_{o}[s]\textbf{R}_{B}[b, t] 
\end{aligned}
\label{Eqn:latching_box}
\end{equation}

Where $\textbf{o}[s]$ is the constant offset vector from the center of the box $b$ to the anchor point $s_b$. $\textbf{R}_{o}[s_b]$ is the constant rotation matrix from the box frame located at the geometric center of the box to the $s_b$ frame located at anchor point $s_b$. This conditional equality constraint can be enforced through big-M formulation such that if $z_{S,i}=1$, ~\eqref{Eqn:latching_box} is enforced.

Each anchor site on the wall $s_{w}$ has two modes. Mode 1 is empty mode where $s_{w}$ is empty, and mode 2 is manipulation mode where $l$ connects to $s_{w}$ as an arm. Multiple $s_{w}$ can exist on the ground. Two binary variables $z_{Wi}[s_{w}, l, t],\text{where } i=1,2$ are used to represent those modes. The associated mode constraints and physical connection constraints are similar to the anchor points on the boxes. We also define binary variables for additional connections between LIMMS as arms or legs: $z_{Ac,i}[l_{pr}, l_{po}, t]$ or $z_{Lc,i}[l_{pr}, l_{po}, t]$, $i=1,2$. If $z_{Ac,i}[l_{pr}, l_{po}, t]=1$, LIMMS $l_{po}$ connects as an additional arm to LIMMS $l_{pr}$ at $t$. Similar for $z_{Lc,i}$.

\begin{figure*}
    \centering
    \includegraphics[width=0.9\textwidth]{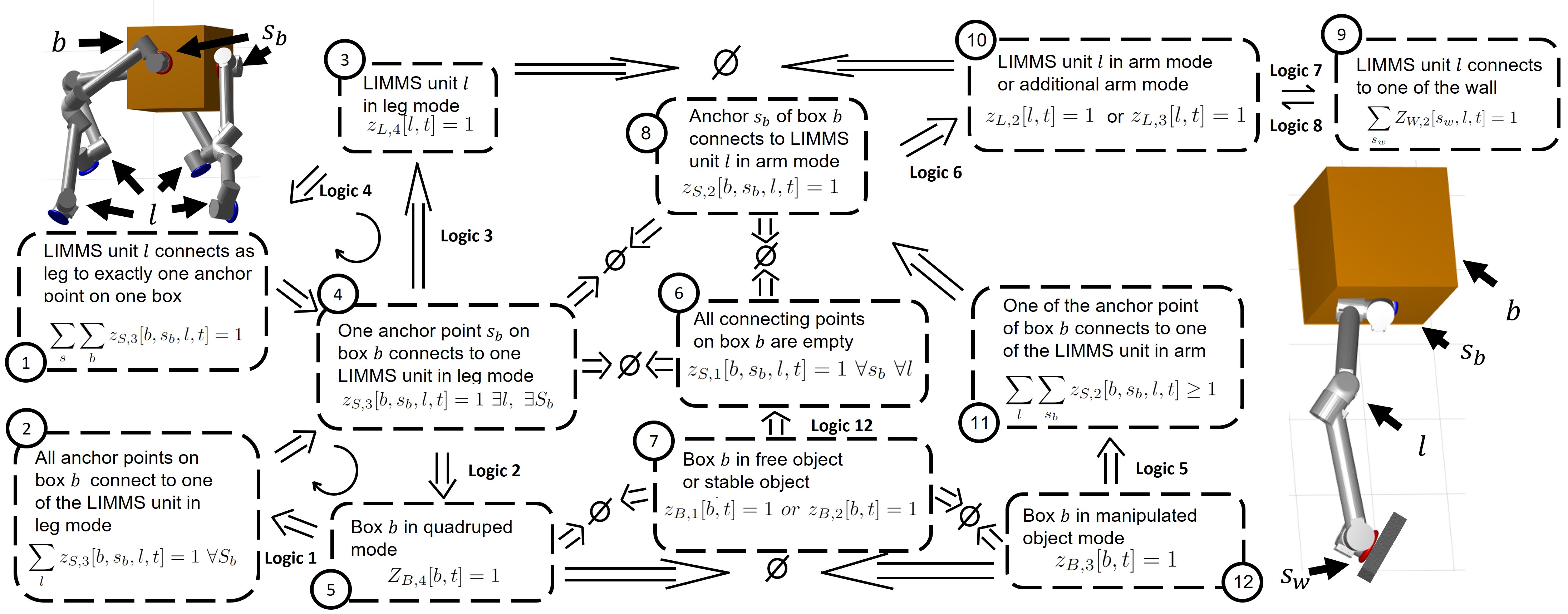} %
    \caption{Implications of logic 1-12. Left half is associated with robot mode while right half is associated with manipulation mode. Arrows are labeled with specific logic rules. If not, the implication is mathematically correct. The symbol $\Rightarrow \emptyset \Leftarrow$ means mutually exclusive.}
    \label{fig:logic}
    \vspace{-0.5cm}
\end{figure*}

\textit{Logic for Boxes as Robot Bodies}. We define 4 logic rules for any box detailed in rule $1-4$ in Table~\ref{tab:logic}. They constrain the boxes in quadruped mode and LIMMS units connected to it.
The gist is to ensure that several things happen simultaneously: 
1) $b$ is used as the robot body, 2) 4 LIMMS are its legs, and 3) connections happen between them. On the other hand, if no box is used as a quadruped body, no LIMMS should be used as legs. This is enforced through formulating the rules into a loop as shown in figure \ref{fig:logic}. If 1 LIMMS is used as leg, it should connect to 1 anchor point on 1 of the boxes due to Logic 4, and the corresponding box should be in quadruped mode due to Logic 2. 

Each logic can be formulated as constraints between integer variables through big-M formulation. For example, Logic $1$ can be written as $1 - M(1-z_{B,4}[b, t]) \leq \sum_{l} z_{S,3}[b, s_{b}, l, t] \leq 1 + M(1-z_{B,4}[b, t])$ where $M$ is a large constant (usually $10^{5}$). Other constraints follow similarly.

\textit{Logic for Boxes as Manipulated Objects}.
We define 7 logic rules for boxes as objects and are manipulated by LIMMS as arms. They again constrain the modes for the box and LIMMS connected to it. The gist is to ensure that the following happens simultaneously: 1) box is being manipulated by 1 or more LIMMS, 2) 1 or more LIMMS operate in arm mode, 3) connections occur between the arm and box, and 4) arm is connected to 1 anchor point on the wall or ground. While most of the logic rules are single directional, the bidirectional rules are enforced by formulating implicit loops. For example, there is no explicit rules enforcing a box to be in manipulated object mode (arrow from $8$ to $12$) if LIMMS arm is manipulating it (block $8$ in Fig. \ref{fig:logic}). However, block $8$ is mutually exclusive with block $4$ and $6$. This means that if $8$ happens, $4$ and $6$ cannot happen, which means block $5$ and $7$ cannot happen (by reversing the direction of implication). This in turn indicates that $12$, which is the complement of $5$ and $7$, has to happen. 
Note that we did not include the constraints for additional arm or leg mode in Fig. \ref{fig:logic}, but similar arguments carry over.

\textit{Logic for Boxes as Stable or Free Objects}. Logic $12$ in Fig. \ref{tab:logic} is defined for boxes as stable or free object modes which states that if a box is stable or free, all its anchors are empty.

\subsection{Continuous Constraints}

\textit{Kinematics}. Kinematics constraints are imposed for each LIMMS through a series of linear constraints and bilinear constraints in the same fashion as \cite{dai2019global}:
\vspace{0.3cm}
\begin{equation}
\begin{aligned}
    & \textbf{p}[j+1, l, t] = \textbf{p}_{L}[j, l, t] + \textbf{R}_{L}[j, l, t]\textbf{p}_{j+1, j} \\
    & \textbf{R}_{L}[j-1, l, t]\textbf{z}_{j-1, j} = \textbf{R}_{L}[j, l, t]\textbf{z}_{j, j} \\
    & \textbf{R}_{L}[j, l, t]\textbf{R}_{L}[j, l, t]^{T}=\textbf{I} \\
    & \textbf{R}_{L}[j, l, t] \ \text{represents a right handed frame}
\end{aligned}    
\label{eqn:kinematics_00}
\end{equation}
\vspace{-0.05cm}

Where $\textbf{p}_{j+1, j}$ is the constant position vector of the next joint as seen in the frame of the previous joint, and $\textbf{z}_{j-1, j}$ is the constant orientation of the next joint as seen in the frame of the previous joint, where ours are $\textbf{z}_{j-1, j} = [0, 0, 1]^{T}$.

\textit{Collision Avoidance with Environment}.
To enforce constraints such that LIMMS and boxes does not collide with the environment, we model the environment into discrete convex regions. All boxes and LIMMS have to stay within the convex regions during the process. We also need to discriminate if the LIMMS or box is making contact with the ground. This introduces additional binary variables $\delta_{Ba, i}$, $\delta_{Bg, i}$, $\delta_{La, i}$, and $\delta_{Lg, i}$. Note subscript $g$ stands for ground, and $a$ stands for air. If LIMMS and boxes are within a convex region, the joint points of LIMMS and corners of boxes are linear combinations of the vertices of the convex region:

\begin{equation}
    \textbf{p} = \sum_{v} \lambda_{v} \textbf{V}_{v}, \ \
    \sum_{v} \lambda_{v} = \delta, \ \ \lambda_{v} \in [0, 1]        
\label{eqn:obstacle_1}
\end{equation}



Where $\textbf{p}$, $\lambda$ and $\delta$ are associated with either corner points of the box $\textbf{c}_{B}$ or position of joints of LIMMS $\textbf{p}_{L}$ as listed in Table~\ref{tab:variables}. $\lambda_{v}$'s represent the vertices of the convex region. If one region is not selected, all $\lambda_{v}$'s are zero due to $\delta=0$.

\textit{Collision Avoidance Between Agents}. To enforce collision avoidance for LIMMS-LIMMS and LIMMS-box contact, we use the formulation from \cite{lin2022reduce} that uses separating planes. For convex polygons, the two polygons do not overlap with each other if and only if there exists a separating hyperplane $\textbf{a}^{T}\textbf{x}=b$ in between \cite{boyd2004convex}. That is, for any point $\textbf{p}_{1}$ inside polygon 1 then $\textbf{a}^{T}\textbf{p}_{1} \leq b$, and for any point $\textbf{p}_{2}$ inside polygon 2 then $\textbf{a}^{T}\textbf{p}_{2} \geq b$. Our problem uses the following constraints:


\begin{equation}
\begin{aligned}
    \textbf{a}^{T}\textbf{p}_{L}[j, l_{1}, t] \leq b, \ \
    &\textbf{a}^{T}\textbf{p}_{L}[j, l_{2}, t] \geq b \\
    \textbf{a}^{T}\textbf{c}_{B}[c, b, t] \leq b, \ \
    \textbf{a}^{T}\textbf{p}_{L}[j, l, t] & \geq b \ \forall t, \ \ \textbf{a}^{T}\textbf{a} \geq 0.5
\end{aligned}
\label{eqn:collision_agent_1}
\end{equation}
\vspace{0.1cm}

    

Where $\textbf{a}$ and $b$ is the normal vector and offset for planes associated with the specific pair. $0.5$ is just an arbitrary nonzero number that we choose, as $\textbf{a}$ does not necessarily need to be a unit vector. With this method, we enforce collision avoidance purely through inequality constraints and avoid using complementary constraints such as \cite{raghunathan2021pyrobocop}.





\textit{Dynamics for Box}. The dynamics are required for the agent to generate strictly feasible motions. However, enforcing dynamics for each LIMMS is expensive given its high DoF. We only enforce dynamics for the boxes. This serves as two purposes. First, it allows the system to generate dynamic motions such as throwing or jumping. Second, it allows the system to select motion plans based on the box weight.

\begin{figure*}[t!]
    \centering
    \includegraphics[width=0.98\textwidth]{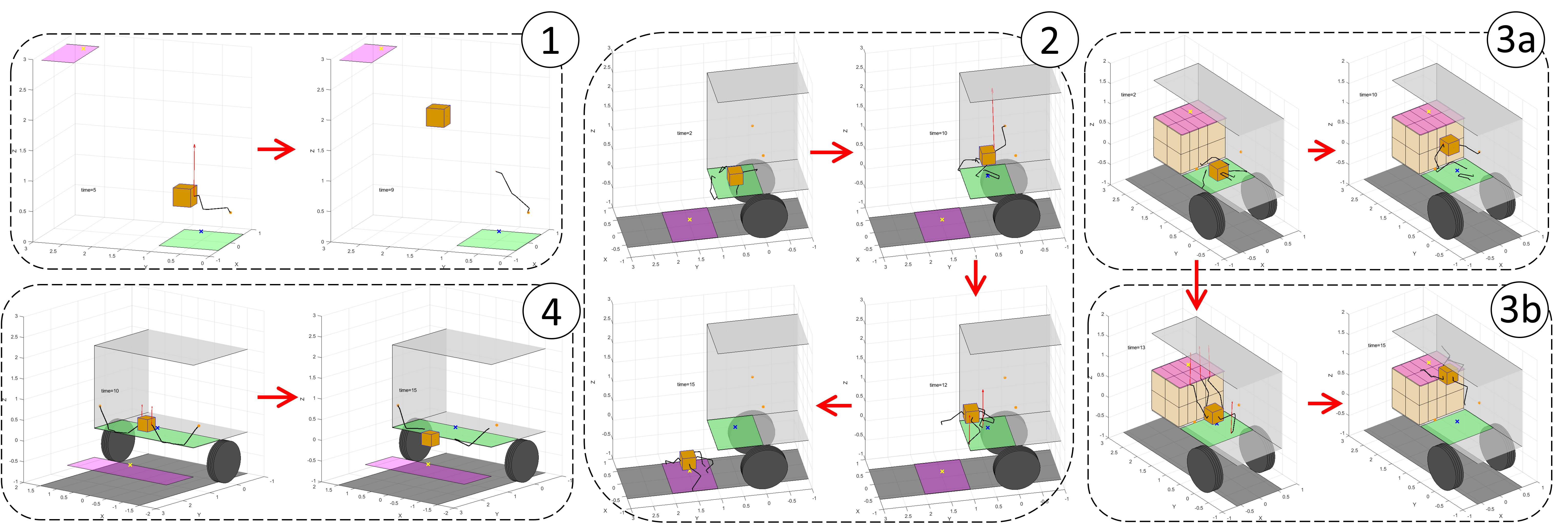} %
    \caption{Results for 5 experiments. The green rectangle represents the initial region in which LIMMS is constrained in. LIMMS is trying to get the box to the magenta region, where LIMMS is allowed to move in as well.: 1) LIMMS picks up a box and throws it towards the goal. 2) LIMMS lifts box in manipulator mode and switches to quadruped mode to jump out of the vehicle. 3a) LIMMS attempts to use manipulator mode to reach goal. 3b) LIMMS climbs onto a step using quadruped mode. 4) LIMMS sends the box towards the goal using dual double arm manipulation to enlarge the workspace.}
    \label{fig:result_figures}
    \vspace{-0.5cm}
\end{figure*}

When the box is in stable object mode, the gravity is compensated for by the ground. Additionally, LIMMS can only apply reaction forces to the box when it is connected through the anchor points on the box. We define the reaction force on the end effector of a LIMMS $l$ to the box $b$ as $\textbf{f}_{i}[b, s, l, t]$, where the index $s$ indicates that the force is through the anchor point $s_b$. When LIMMS connects to the box as a leg, $\textbf{f}_{i}$ serves as the contact force on the ground, while when LIMMS connects to the box as an arm, $\textbf{f}_{i}$ serves as the contact force to grasp the box. The box dynamics are: 
\begin{equation}
    m \ddot{\textbf{p}}_{box}[b, t] = \sum_{s=1}^{4} \sum_{l=1}^{L} \textbf{f}_{i}[b, s, l, t] - m\textbf{g}(1-z_{B,1}[b, t])
\label{eqn:linear_dynamics}
\end{equation}

\begin{equation}
    \sum_{s=1}^{4} \sum_{l=1}^{L} (\textbf{p}_{L}[j=6, l, t] - \textbf{p}_{B}[b, t]) \times \textbf{f}_{i}[b, s, l, t] = 0
\label{eqn:rotational_dynamics}
\end{equation}

$\textbf{f}_{i}[b, s, l, t]$ only exists when $l$ is connected to the box:
\vspace{0.3cm}
\begin{equation}
    \textbf{f}_{i}[b, s, l, t]=0 \ \ \text{if} \ \ z_{S,1}[b, s_{b}, l, t] = 1
\label{eqn:force_shutdown}
\end{equation}

When $l$ connects to the box as leg mode, $\textbf{f}$ represents the contact force from the ground. Therefore, $\textbf{f}$ needs to satisfy the friction cone constraint:
\vspace{-0.000001cm}
\begin{equation}
    \textbf{f} \in Cone \ \ \text{if} \ \ z_{S,i}[b, s_{b}, l, t] = 1 \ \text{and} \ \delta_{Lg}[j=6, l, \exists p, t]
\label{eqn:friction_cone}
\end{equation}

Since LIMMS has a loading capacity, we enforce the max norm constraint on any contact force: $||\textbf{f}|| \leq f_{max}$.
Note that there is a contact moment across the latching. Missing this moment results in the box incapable of being manipulated with a single contact (moment balance will be violated). For simplicity, we fix the box orientation: $\textbf{R}=\textbf{I}$. This can be justified as in manipulated mode, the contact moment is sufficient to keep the orientation of the box, and in quadruped mode, the four ground support points is sufficient to keep the body orientation.

\textit{Support Polygon}. The balance of moment constraint should guarantee the stability of the quadruped body. However, there are already many nonlinear constraints such as kinematics. To simplify the NLP formulation, one approach is to enforce a simple stability constraint in replacement of the moment balance constraint. Since the rotation matrix is fixed, we can simply enforce that the end effector of LIMMS stays to one side of the body which guarantees that the body's center of mass lies within the support polygon of the foot.

\textit{Stability of LIMMS}. One drawback of our formulation is that we do not include the dynamics of LIMMS. As a compensation, there should be a constraint to guarantee the stability when LIMMS is on the ground and tries to reach an anchor point. 
We just assume that we have many latching points on the ground so realizing this motion is relatively simple. Therefore, when LIMMS is in free mode, we enforce that the base stays on the ground and give a speed constraint:

\begin{equation}
    ||\textbf{p}_{L}[j, l, t+1] - \textbf{p}_{L}[j, l, t]|| \leq \Delta \textbf{P} \ \ \text{if} \ \ \delta_{Lg}[j=0, l, p, t] = 1
    \label{eqn:stability}
\end{equation}

\textit{Continuity of Connection}. One suboptimal solution to avoid is having LIMMS frequently latch on and off an anchor point. We enforce equality constraints for binary variables within a range $z_{S,i}[t] = ... = z_{S,i}[t+n]$. We usually choose $n$ between $3$ and $5$, decided based on the speed of latching. 


%% file: Sections/S4_solving_with_ADMM.tex
\section{ADMM Formulation}
\label{sec:ADMM}

Collecting the constraints defined previously, the problem to solve becomes:

\begin{equation}
\begin{aligned}
& \underset{z, \ \delta, \ \textbf{p}, \ \textbf{R}, \ \textbf{f}, \ \lambda, \ \textbf{a}, \ b}{\text{minimize}} \quad f_{obj} \\
& \text{subject to} \\
& \text{\textbf{Mixed integer constraints:}} \\
& \text{Logic rules 1-12} \\
& \text{collision avoidance with the environment ~\eqref{eqn:obstacle_1}} \\ 
& \text{Dynamics constraints: ~\eqref{eqn:linear_dynamics}, ~\eqref{eqn:force_shutdown}, ~\eqref{eqn:friction_cone}} \\
& \text{Stability of LIMMS on ground ~\eqref{eqn:stability}}\\
& \text{\textbf{Nonlinear constraints:}} \\  
& \text{Kinematics ~\eqref{eqn:kinematics_00}} \\
& \text{Collision avoidance between agents ~\eqref{eqn:collision_agent_1}} \\
& \text{Dynamics constraints ~\eqref{eqn:rotational_dynamics}}
\end{aligned}
\label{eqn:problem_1}
\end{equation}

Where $f_{obj}$ is a quadratic equation that minimizes the distance of the box to the goal position. The variables and constraints of problem ~\eqref{eqn:problem_1} incorporate discrete and continuous variables with linear and nonconvex (bilinear) constraints. This results in an MINLP. The commercial solvers tend to perform non-satisfactory in this type of problem. There are generally two approaches that convert this type of problem: an MICP using convex envelope relaxations for nonlinear constraints \cite{dai2019global} or conversion of the discrete variables into continuous ones through complementary formulation \cite{stein2004continuous}. MIPs with convex envelopes tend to solve slowly when the problem scales up. Since there are many discrete variables in this problem, complementary formulations will be numerically difficult \cite{shirai2022simultaneous}. In this paper, we adopt the ADMM. ADMM separates the problem into two sub-problems. Although those sub-problems have different constraints, ADMM iterates between sub-problems such that constraint $1$ which may not appear in sub-problem $2$ will be implicitly enforced as the iteration proceeds. In the end, sub-problems will reach a consensus meaning their solutions are close to each other. This procedure is detailed in Algorithm \ref{alg:ADMM}. In our problem, the logic rule constraints are resolved through MIPs, while the nonlinear kinematics and collision avoidance constraints are resolved through NLPs. Similar to \cite{aydinoglu2021real}, we first make copies $\textbf{var}_{2}$ of the variables $\textbf{var}_{1}=[z, \delta, \textbf{p}, \textbf{R}, \textbf{f}, \lambda, \textbf{a}, b]$. Represent the feasible set of mixed-integer constraints through $0-\infty$ indicator function by $\mathcal{I}_{M}$ and the nonlinear constraints by $\mathcal{I}_{N}$. The consensus problem between MIP and NLP is:
\begin{equation}
\begin{aligned}
& \underset{\textbf{var}_{1} \ \textbf{var}_{2}}{\text{minimize} \ \ } f_{obj} + \mathcal{I}_{M}(\textbf{var}_{1}) + \mathcal{I}_{N}(\textbf{var}_{2}) \\
& s.t. \ \ \textbf{var}_{1} = \textbf{var}_{2}
\end{aligned}
\label{eqn:problem_2}
\end{equation}
\vspace{-0.000001cm}
The constraints are moved to the objective function through the indicator function. Applying ADMM \cite{boyd2011distributed} to the Lagrangian $\mathcal{L}$ of ~\eqref{eqn:problem_2} results in three iterative operations:
\begin{subequations}
\begin{flalign}
& \textbf{var}_{1}^{i+1} = \text{argmin}_{\textbf{var}_{1}} \mathcal{L}(\textbf{var}_{1}^{i}, \ \textbf{var}_{2}^{i}, \ \textbf{w}^{i}) \label{eqn:MIP_step} \\ 
& \textbf{var}_{2}^{i+1} = \text{argmin}_{\textbf{var}_{2}} \mathcal{L}(\textbf{var}_{1}^{i+1}, \ \textbf{var}_{2}^{i}, \ \textbf{w}^{i}) \label{eqn:NLP_step} \\
& \textbf{w}^{i+1} = \textbf{w}^{i} + \textbf{var}_{1}^{i} - \textbf{var}_{1}^{i} \label{eqn:Dual_update}
\end{flalign}
\label{eqn:ADMM}
\end{subequations}
Where $\textbf{w}$ is the dual variable of the Lagrangian of ~\eqref{eqn:problem_2}. In ~\eqref{eqn:ADMM}, ~\eqref{eqn:MIP_step} solves the MIP problem:
\begin{equation*}
\begin{aligned}
& \underset{\textbf{var}_{1}}{\text{minimize}} \quad ||\textbf{var}_{1}^{i}-\textbf{var}_{2}^{i}+\textbf{w}^{i}||_{\textbf{W}_{\text{MIP}}^{k}} \\
& \text{s.t.  Mixed-integer constraints in ~\eqref{eqn:problem_1}} 
\end{aligned}
\end{equation*}

In the next step, ~\eqref{eqn:NLP_step}, solves the NLP:
\begin{equation*}
\begin{aligned}
& \underset{\textbf{var}_{2}}{\text{minimize}} \quad ||\textbf{var}_{2}^{i}-(\textbf{var}_{1}^{i+1}+\textbf{w}^{i}))||_{\textbf{W}_{\text{NLP}}^{k}} \\
& \text{s.t.  Nonlinear constraints in ~\eqref{eqn:problem_1}} 
\end{aligned}
\end{equation*}

And the next step, ~\eqref{eqn:Dual_update}, updates the dual variable $\textbf{w}$. To finish one iteration, the weights for MIP, $W_{\text{MIP}}$, the weights for NLP, $W_{\text{NLP}}$, and the dual variable $\textbf{w}$, are updated with line $6-7$ in Algorithm \ref{alg:ADMM}. Within one iteration, ~\eqref{eqn:MIP_step}, ~\eqref{eqn:NLP_step}, ~\eqref{eqn:Dual_update} are solved in succession. This iterative procedure continues until the discrepancy between the MIP solutions and the NLP solutions $\boldsymbol{\theta} = \textbf{var}_{1}^{i} - \textbf{var}_{1}^{i}$ are lower than the user-set error threshold $\boldsymbol{\theta}_{th}$.

It is well known that ADMM has convergence guarantees for convex problems and can significantly improve the solving speed. However, for complex MINLPs, there is no convergence guarantee. In this problem, both MIP and NLP can be slow and expensive. We avoid explicitly placing complementary constraints to represent discrete modes as \cite{aydinoglu2021real} did, since it hinders convergence for NLP. However, NLP does need some information on discrete variables as it needs to reason connections and turn variables such as $\textbf{f}$ on or off accordingly. After solving the MIPs, we directly use the solutions of $z_{S,i}$, $z_{W,i}$ in the NLP step to enforce connections. This improves the precision of the NLP step and the overall precision of consensus. The price to pay is an increase in difficulty for NLP solvers to find solutions.

\begin{algorithm}[t] 
\small 
\algsetup{linenosize=\small}
\caption{$\operatorname{ADMM \ for \ LIMMS}$}
\label{alg:ADMM}
\hbox{\textbf{Input} $\rho$, $\textbf{W}_{\text{MIP}}^{0}$, $\textbf{W}_{\text{NLP}}^{0}$, $\textbf{w}^{0}$, $\textbf{var}_{2}^{0}$, $\boldsymbol{\theta}_{th}$}
\begin{algorithmic}[1] \label{alg:envfit}
\STATE Initialization $i=1$
\WHILE{\text{$\boldsymbol{\theta} > \boldsymbol{\theta}_{th}$ \textbf{and} $i < i_{max}$}}
\STATE Compute $\textbf{var}_{1}^{i+1}$ via $~\eqref{eqn:MIP_step}$
\STATE Compute $\textbf{var}_{2}^{i+1}$ via $~\eqref{eqn:NLP_step}$
\STATE $\textbf{w}^{i+1} \longleftarrow \textbf{w}^{i} + \textbf{var}_{1}^{i} - \textbf{var}_{1}^{i}$
\STATE $\textbf{W}_{\text{MIP}}^{k+1} \longleftarrow \rho \textbf{W}_{\text{MIP}}^{k}$, \ \ $\textbf{W}_{\text{NLP}}^{k+1} \longleftarrow \rho \textbf{W}_{\text{NLP}}^{k}$ 
\STATE $\textbf{w}^{i+1} \longleftarrow \textbf{w}^{i} / \rho$
\STATE $\boldsymbol{\theta} \longleftarrow \textbf{var}_{1}^{i} - \textbf{var}_{1}^{i}$
\STATE $i=i+1$
\ENDWHILE
\RETURN $\textbf{var}_{2}^{i}$
\end{algorithmic} 
\end{algorithm}

%% file: Sections/S5_Results.tex
\section{Results}
\label{sec:results}
\begin{figure}[t!]
    \centering
    \includegraphics[width=0.5\textwidth]{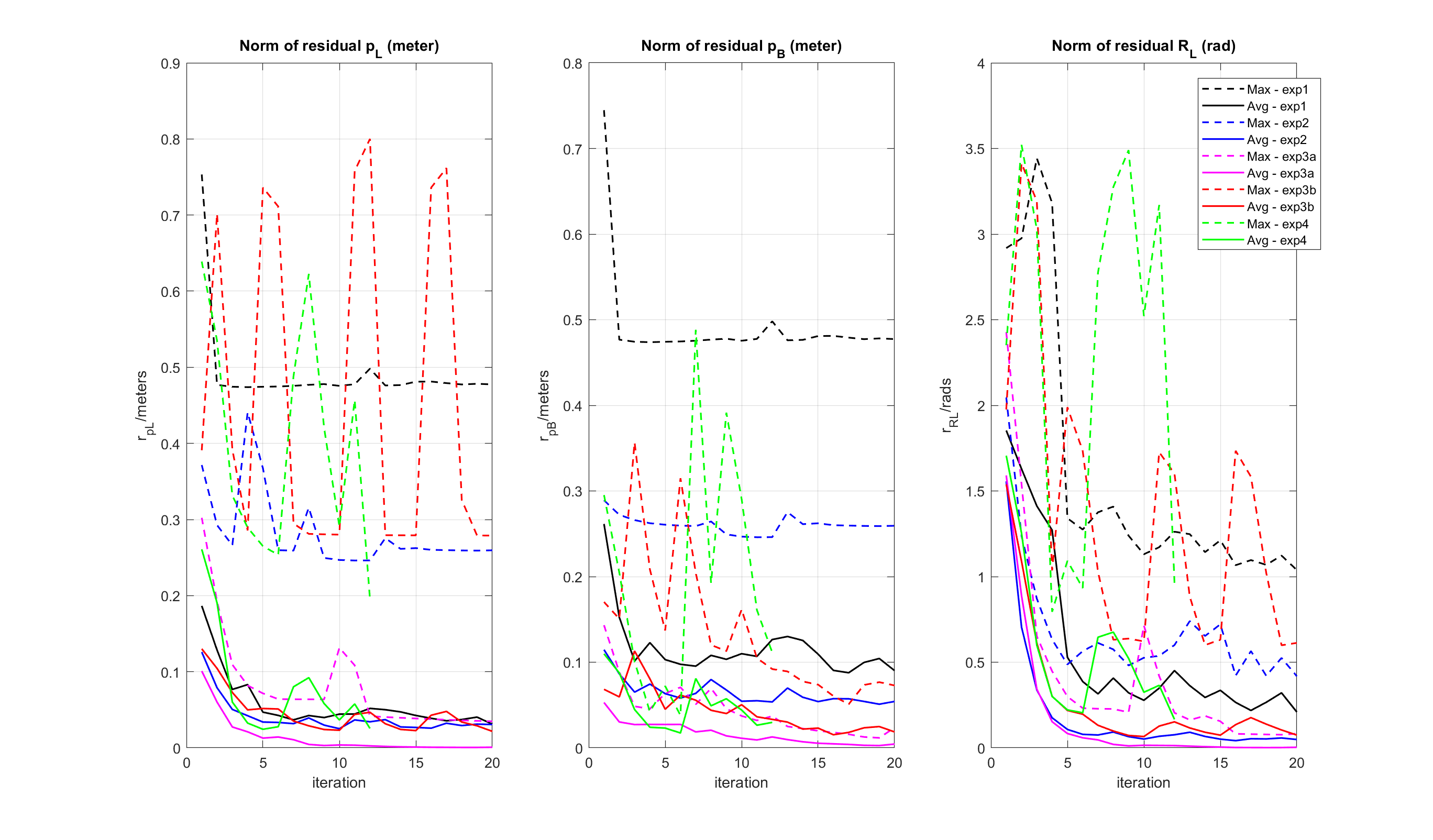} %
    \caption{Convergence of mean and max residual $r_{pB}$, $ r_{pL}$, $ r_{RL}$ for experiment 1-4. Solid lines denote mean residuals and dashed lines denote max residuals.}
    \label{fig:error_convergence}
\end{figure}

We performed 5 numerical experiments to evaluate the performance of the proposed formulation and ADMM algorithm. For all experiments, the MIP formulation was solved with Gurobi 9, and the NLP formulation was solved through Ipopt on a Intel Core i7-7800X 3.5GHz $\times$ 12 machine. We solved all the scenarios for 15 iterations. Other parameters we have chosen are: $\Delta t=1 sec$, $f_{max}=20N$, max moving speed for LIMMS on the ground of $0.5 m/s$, and a $30cm$ cube box. The 6 Dof LIMMS unit lengths are $5cm$, $7.4cm$, $33cm$, $7.8cm$, $33cm$, $7.4cm$, $5cm$, respectively for each link starting at one end. Fig. \ref{fig:result_figures} shows the MATLAB visualization of the first 4 experiments. Animations of all tests are included in the attached video available at \url{https://www.youtube.com/watch?v=RH9gMOK24L0}.

1) \textit{Throwing}. In the first experiment, we placed 1 LIMMS and 1 box in the scene and provided 1 anchor point on the wall. The goal is set higher than a LIMMS unit can reach. The solver gives a solution where the LIMMS unit connects to the wall and picks up the box, then throws the box to the goal. This simple test shows the ability of the solver to generate manipulation motions using dynamics. 

2) \textit{Jumping}. In the second experiment, we placed 4 LIMMS and 1 box on a raised platform or inside a truck and set the goal position to be lower on the ground. In addition, 2 anchor points are provided on the wall. The planner provides the solution where 1 LIMMS latches itself onto the wall and lifts the box. Then the other 3 LIMMS connects to the anchor points on the sides of the box which enters quadruped mode. This quadruped robot then jumps down the step to reach the goal. The solver automatically changes modes from manipulation to quadruped.

3a,b) \textit{Weight Lifting}. In the third experiment, we investigate the behavior change due to a change in the weight of the box. We place 4 LIMMS and 1 box down a ledge and set the goal to be above the ledge. Two anchor points on the wall are provided. First, we set the box weight to be 0.5kg. The planner provides a manipulation trajectory where two LIMMS connect to the wall and lift the box to move it to the goal. We then increase the box weight to 7kg. In this case, if we force LIMMS to manipulate the object, the planner returns infeasible. The feasible trajectory returned by the solver set 3 LIMMS to be anchored onto the ledge and box, while 1 leg stays down from the ledge to push the body up onto the platform. As the box weight exceeds the capacity of dual arm manipulation, a quadruped motion is necessary to lift the 7kg box.

4) \textit{Manipulation with Double-dual Arms}.
In the fourth experiment, we again place 4 LIMMS and 1 box in the same scenario, below a desired platform or step. Two anchor points are provided on the wall. However, we enlarged the width of the truck such that 1 LIMMS cannot reach the box from the wall. In this case, the solver connected a second LIMMS as an additional arm to the first which is connected to the wall, allowing the arms to reach the box and perform dual arm manipulation.

5) \textit{Quadruped Walking with Refinement}. In the fifth experiment, we placed 4 LIMMS and a box on flat ground and set the goal to be at the same elevation but separated by a distance. LIMMS moved to the box and assembled a quadruped robot which then moved towards the goal. Furthermore, since there is no gait optimization in our problem formulation, we used an additional planner similar to \cite{bledt2019implementing} but included 6 DoF kinematics to provide the quadruped walking motion to the goal. This experiment demonstrated that, although the planner may give rough trajectories, they can be further refined with another planner to correct kinematics discrepancies and gait cycles. This readies it to be implemented on the hardware.

6) \textit{Box Lifting on Hardware} In this experiment, one LIMMS is initially anchored to the ground. There is another anchor point on the wall. The objective is to lift the box higher. If LIMMS lifts the box with the other end anchored to the ground, the kinematics quickly becomes infeasible. The planner instead lets LIMMS first anchor to the box, then anchor to the wall from the box. The LIMMS can easily lift the box higher with the other end anchored to the wall. The hardware implementation is included in the video.

7) \textit{Convergence} Define the residual to be the mismatch between the MIP and NLP solutions. The mean residuals for $\textbf{p}_{B}$, $\textbf{p}_{L}$ and $\textbf{R}_{L}$ are the mean value of all the norms:
\begin{subequations}
\begin{flalign}
    & r_{pB}[i] = \underset{b}{\text{mean}} \ ||\textbf{p}_{B,\text{MIP}}[b, i] - \textbf{p}_{B,\text{NLP}}[b, i]|| \\
    & r_{pL}[i] = \underset{j, \ l}{\text{mean}} \ ||\textbf{p}_{L,\text{MIP}}[j, l, i] - \textbf{p}_{L,\text{NLP}}[j, l, i]|| \\
    & r_{RL}[i] = \underset{j, \ l}{\text{mean}} \ ||\text{Vec}(\textbf{R}_{L,\text{MIP}}[j, l, i] - \textbf{R}_{L,\text{NLP}}[j, l, i])||
\end{flalign}
\end{subequations}
The max residual is the maximal value of all the norms. Figure \ref{fig:error_convergence} depicts the change of mean and max residuals as a function of time. ADMM generally showed decent average consensus after iteration 10, where $r_{pB}$ or $r_{pL}$ usually converges to cm-mm level and $r_{RL}$ usually less than 0.1 rad. The maximal residual can sometimes be large. If we put the MIP solutions on the real hardware, we can run another kinematics refiner to solve the kinematics to ensure that the nonlinear constraints are strictly satisfied.

The number of variables, constraints and time cost for solving the experiments above are listed in Table~\ref{tabl:time_cost}. Generally, the NLP portion is the more challenging portion of ADMM, since it includes kinematic and collision avoidance constraints. To speed up the solving process, some linear constraints in \label{eqn:kinematics} can be moved into the MIP formulations. This will speed up the NLP solver but the residual may increase.

\begin{table}[]
\centering
\setlength{\tabcolsep}{2.7pt}
\begin{tabular}{lllrrlrrr}
\hline
\multicolumn{1}{l}{\multirow{2}{*}{\#}} & \multirow{2}{*}{T} & \multicolumn{2}{c}{\# of variables}                                    & \multicolumn{2}{c}{\# of constraints} &  \multicolumn{3}{c}{Time in Minutes}\\ \cline{3-9}
\multicolumn{1}{c}{}                      &                    & MIP                                                             & NLP  & MIP   & NLP                                                            & T-MIP                       & T-NLP                       & T-Total                         \\ \hline
1                                  & 10                 & \begin{tabular}[c]{@{}l@{}}4857 cont.\\ 670 bin.\end{tabular}   & 1777 & 3623  & \begin{tabular}[c]{@{}l@{}}eq. 582\\ ineq. 5206\end{tabular}   & 0.88                                     & 7.17                                    & 8.05                                       \\ \hline
2                                  & 15                 & \begin{tabular}[c]{@{}l@{}}25917 cont.\\ 3150 bin.\end{tabular} & 8787 & 34545 & \begin{tabular}[c]{@{}l@{}}eq. 621\\ ineq. 28990\end{tabular}  & 129                                     & 24                                      & 153                                       \\ \hline
3a)                                 & 15                 & \begin{tabular}[c]{@{}l@{}}25917 cont.\\ 3150 bin.\end{tabular} & 8787 & 33102 & \begin{tabular}[c]{@{}l@{}}eq. 1686\\ ineq. 28990\end{tabular} & 41                                      & 107                                     & 148                                       \\ \hline
3b)                                   & 10                 & \begin{tabular}[c]{@{}l@{}}17277 cont.\\ 2100 bin.\end{tabular} & 5857 & 22032 & \begin{tabular}[c]{@{}l@{}}eq. 1131\\ ineq. 19180\end{tabular} & 5                                       & 50                                      & 55                                        \\ \hline
4                               & 15                 & \begin{tabular}[c]{@{}l@{}}25917 cont.\\ 3690 bin.\end{tabular} & 8787 & 35924 & \begin{tabular}[c]{@{}l@{}}eq. 606\\ ineq. 31342\end{tabular}  & 19                                      & 244                                     & 263                                       \\ \hline
\end{tabular}
\caption{Solving time for experiment 1-4. Note: Far left column in ascending order is: 1 for Throw, 2 for Jump, 3a for Lifting with Arm, 3b for Lifting with Climb, and 4 for Double-dual Arm.}
\label{tabl:time_cost}
\end{table}

%% file: Sections/S6_Conclusion.tex
\section{Conclusion and Future Work}
\label{sec:conclusion}
\vspace{-0.1cm}
This paper presents an optimization based motion planner for the multi-agent modular robot system LIMMS. We demonstrated solving the proposed formulation with ADMM. The results show how LIMMS autonomously coordinates between different modes and generates trajectories of the system under different situations. With proper refinement, the trajectories can be implemented on the hardware.

It is worthwhile to mention that due to the separating plane for collision avoidance, the NLP part of the problem takes a long time to solve. In the previous work \cite{lin2022reduce}, we used data-driven methods to solve those constraints fast online. A similar approach can be adopted here for future work.

Future work includes more testing of the ADMM design. Since the NLP portion of our implementation of ADMM can be slow and sometimes difficult to converge, further decompositions of the NLP to smaller problems may be helpful. Additionally, some of the logic variables may be simplified. For example, the mode of the box can be blended such that the box can have arms and legs simultaneously. Other future work include development of a coordinated control algorithm for larger scale implementations of LIMMS.